\RequirePackage{fix-cm}
\documentclass[twocolumn]{svjour3}          
\smartqed  
\usepackage{graphicx}
\usepackage{latexsym}
\usepackage{url}
\usepackage{caption}
\usepackage{xcolor}
\usepackage{mathtools}
\usepackage{mathptmx}      
\usepackage{array}
\usepackage[numbers]{natbib}

\makeatletter
\newcommand{\thickhline}{%
    \noalign {\ifnum 0=`}\fi \hrule height 1pt
    \futurelet \reserved@a \@xhline
}
\newcolumntype{"}{@{\hskip\tabcolsep\vrule width 1pt\hskip\tabcolsep}}
\makeatother
\begin{document}

\title{Cluster validity index based on Jeffrey divergence
}


\author{Ahmed Ben Said      \and
        Rachid Hadjidj \and Sebti Foufou
}


\institute{Ahmed Ben Said \at
              CSE Department, College of Engineering, Qatar University, P.O. Box 2713, Doha-Qatar  \\
							LE2i Lab, UMR CNRS 6306, University of Burgundy, BP 47870, 21078, Dijon-France
              \email{abensaid@qu.edu.qa}           
           \and
           Rachid Hadjidj . Sebti Foufou \at
              CSE Department, College of Engineering, Qatar University,  P.O. Box 2713, Doha-Qatar
}

\date{Received: date / Accepted: date}

\maketitle

\begin{abstract}
Cluster validity indexes are very important tools designed for two purposes: comparing the performance of clustering algorithms and determining the number of clusters that best fits the data. These indexes are in general constructed by combining a measure of compactness and a measure of separation. A classical measure of compactness is the variance. As for separation, the distance between cluster centers is used. However, such  a distance doesn't always reflect the quality of the partition between clusters and sometimes gives misleading results. In this paper, we propose a new cluster validity index for which Jeffrey divergence is used to measure separation between clusters. Experimental results are conducted using different types of data and comparison with widely used cluster validity indexes demonstrates the outperformance of the proposed index.
\keywords{Clustering \and cluster validity index \and Jeffrey divergence}
\end{abstract}

\section {Introduction}
Cluster analysis refers to the procedures by which hidden groups in data are unveiled \cite{eltit}.
Several clustering techniques have been proposed in the literature  and can be divided into three categories: hierarchical, density based, and partitional \cite{tran2005a}. Hierarchical clustering consists of successive clustering operations until a desired partition is obtained.  The technique starts by considering each pixel as a cluster then similarities between pairs of clusters are computed and the most similar clusters are merged. These operations will lead to a diagram called \emph{ dendrogram} \cite{tran2005a} representing  nested clusters. To obtain a specific partition, the dendrogram is cut at a given level. In density based clustering, clusters are considered as regions. Features belonging to every region are dense, then the density is estimated around these features. Each local maximum represents a cluster. The best known density-based clustering method is DBSCAN \cite{Ester96adensitybased}. Partitional clustering seeks to divide data \textit{X}={$ x_{1} $,$ x_{2} $,\ldots,$ x_{n} $} -where $ x_{i} $ is a $\textit{d}$-dimensional data- into a given number \textit{k} of clusters: $ C_{1} $,$ C_{2} $,\ldots,$ C_{k} $ \cite{krooshof2012}. This procedure is conducted by minimizing an objective function. The best known partitional clustering is the K-means algorithm \cite{Jain2010651}. K-means minimizes the sum of squared distances between features and cluster centers ($ v_{1} $,$ v_{2} $,\ldots,$ v_{k} $). The cost function of K-means is:
\begin{equation}
 J = \sum_{i=1}^k  \sum_{x_{j}\in C_{i}} ||x_{j}-v_{i}||^{2} 
\end{equation}
A fuzzy version of K-means called the fuzzy C-means (FCM) was proposed by  Bezdek \cite{bezdek1981} where each $ x_{j} $ data objects belongs to every cluster $ v_{i} $ with a degree called the membership degree $ u_{ij}  \in  {[}0,1{]} $ of the $j^{th}$ data object in the $i^{th}$ cluster. n is the number of data objects. The fuzzy objective function $ J_{FCM}$ is:
\begin{equation}
J_{FCM}=\sum_{i=1}^k  \sum_{j=1}^n u_{ij}^{m} \cdot ||x_{j}-v_{i}||^{2} ;  \hspace{1cm} 1  <  m  <  \infty
\label{eq2}
\end{equation}
The aforementioned algorithms, similarly to various partitional algorithms, require a predefined number of clusters. Cluster validity indexes (CVIs)  are used to look for the optimal number of clusters that best fit the data, in addition to their ability to  discriminate between clustering algorithms \cite{Pakhira2005191}. CVIs are computed after applying a clustering algorithm. Many CVIs are  proposed in the literature \cite{weina2007} and several clustering problems have been addressed such as the presence of cluster with different sizes and densities \cite{Pascual2010454,Rizman} and the presence of noise and outliers \cite{Wu20092541}. These indexes are generally constructed  by combining a measure of cluster compactness  and a measure of separation between clusters. Compactness refers to the degree of closeness of the data points to each other. A classical measure of compactness is the variance which is the average of the squared distances to the mean. A low variance value is an indicator of a good partition. The separation reflects how clusters are separated from each others. It is generally determined by measuring the distance between cluster centers. High separation value indicates a good partition quality. \newline Many studies have focused on CVIs comparison and performance evaluation and almost all of them have opted for the following method: a clustering algorithm is run over data sets with  known number of clusters that best fits the data.  We compute CVIs for each clustering while varying the number of clusters. The CVI that points to the known number of clusters is the best \cite{weina2007}. A recent alternative methodology is proposed and tested for CVIs evaluation \cite{Arbelaitz2013243,Gurrutxaga2011505} in which the definition of the optimal partition has been questioned. Indeed, this partition is now defined as the most similar partition to the perfect partition. Thus, if the CVI points to this particular partition, then it is the best.  CVIs are roughly divided into two categories: CVIs based only on membership value and CVIs based on data set besides the membership value. \newline In this work, we propose a new CVI based on a new separation measure. The classical separation measure which uses distances between features and cluster centers does not reflect the real separation especially when it comes to clusters with different sizes and densities. One of the possible solutions is to design a distance metric that is adapted to each data point as proposed in \cite{xing} to ensure a significant separation measure. This unsupervised distance metric learning approach aims at incorporating the maximum of discriminative information to design a metric that is able to regroup similar data samples in one class and dissimilar samples in different classes. The learning process  can be global or local. With a global learning, distance between pairs is minimized according to the equivalence constraints. Separation of the data pairs is conducted using the inequivalence constraints. However, in case of classes that exhibit multimodal distributions, a conflict between theses equivalence and inequivalence constraints may occur. In \cite{Mu20132337}, authors proposed a local discriminative distance metrics algorithm that is not only capable of dealing with the problem of the global approach but also incorporate multiple distance metrics unlike the previous works \cite{Sugiyama,Weinberger} where a single distance metric is learned on all the data sets. Our approach is based on considering each cluster as a density to be estimated. Thus  we suggest a new separation measure between clusters based on Jeffrey divergence \cite{distance} between clusters. Comparisons with various CVIs are conducted on several synthetic and real data sets.\newline \indent The rest  of this document is organized as follows: in section 2, we give an overview of well-known CVIs. In section 3, we present the motivation behind this work and detail the suggested CVI. Section 4 is dedicated to the experimentation where comparison with various CVIs is conducted. We conclude in section 5.  
\vspace{-0.4cm}
\section {Cluster validity index}
Many cluster validity indexes have been proposed in the literature and extensive studies have been conducted to evaluate their performances.  Next we describe widely used CVIs.\vspace{-0.4cm} 
\subsection{CVIs based on membership value}
Bezdek proposed a cluster validity index called the partition coefficient \textit{(PC)} based on membership values \cite{bezdek1981}. The index is computed as the sum of squared membership values of the obtained partition divided by the number of features. This  measures the amount of membership sharing between clusters and ranges in {[}1/c,1{]}. The optimal partition is the one with maximum  \textit{PC}.
\begin{equation}
PC=\frac{1}{n}\sum_{i=1}^c  \sum_{j=1}^n u_{ij}^{2}
\label{eq3}
\end{equation}
Bezdek proposed also another CVI based on membership values called the partition entropy \textit{PE} \cite{Bezdek1974}. \textit{PE} measures the amount of fuzziness of the obtained partition. The best partition is the one with the least \textit{PE} value.
\begin{equation}
PE=- \frac{1}{n} \sum_{i=1}^k  \sum_{j=1}^n u_{ij} \cdot log(u_{ij})
\label{eq4}
\end{equation}
Cluster validity index \textit{P} \cite{Chen2004} combines a measure of intracluster compactness and intercluster separation based only on membership values. The maximal value of \textit{P} points to the best partition.
\begin{equation}
P=\frac{1}{n} \sum_{j=1}^{n} \underset{i} {max}(u_{ij}) - \frac{1}{K} \sum_{i=1}^{k-1} \sum_{j=i+1}^{k} \left[ \frac{1}{n} \sum_{l=1}^{n} min (u_{il},u_{jl}) \right]
\label{eq5}
\end{equation}
\noindent
With $ K= \sum_{i=1}^{k-1} i $. \newline
In the first term, the closer a feature $x_{j}$ to the $i^{th}$ cluster, the closer $ \underset{i}{max}(u_{ij}) $ is to 1. In the second term, the closer a feature $x_{j}$ to the $i^{th}$ cluster, the closer min$(u_{il},u_{jl}) $ is to 0 meaning that cluster $i$ and $j$ are well separated. On the other hand, if min$(u_{il},u_{jl}) $ is close to 1/c, then $x_{l}$ equally belongs to cluster $i$ and $j$ and we have the fuzziest partition. 
The aforementioned CVIs have some drawbacks such as their monotonic dependency on the number of clusters and the absence of data features itselves. \vspace{-0.4cm} 
\subsection{CVIs based on membership value and data set}
Xie \cite{xie} proposed a cluster validity index called  $XB$.  The index is computed as the ratio of the within cluster compactness to cluster separation computed as the minimum distance between cluster centers.
\begin{equation}
XB=\frac{\sum_{i=1}^k  \sum_{j=1}^n u_{ij}^{2} \cdot ||x_{j}-v_{i}||^{2}}{n \cdot \underset{i \neq l}{min} ||v_{i}-v_{l}||^{2}}
\label{eq6}
\end{equation}
Pakhira \cite{Pakhira2004487} proposed another CVI that mixes compactness and separation called the $PBMF$ index.
\begin{equation}
PBMF= \left(\frac{1}{c} \cdot \frac{E_{1}}{J_{m}} \cdot D_{c}\right)^{2}
\label{eq7}
\end{equation}
with $J_{m}=\sum_{i=1}^k  \sum_{j=1}^n u_{ij} \cdot ||x_{j}-v_{i}||$ , $E_{1}=\sum_{j=1}^n ||x_{j}-v_{1}||$ and $D_{c}= \underset{1\leq i,j\leq k}{max} ||v_{i}-v_{j}||$ 
\newline \newline
\noindent A modified version of the $PBMF$ index  called PBM-FVG is introduced by including a factor called the granulation error \cite{Bandyopadhyay201122}. The granulation error is caused by the clustering algorithm.  Let $\hat{x_{j}}$ be the estimate of $x_{j}$.
\begin{equation}
\hat{x_{j}}=\frac{\sum_{i=1}^k u_{ij}^{2} \cdot v_{i}}{\sum_{i=1}^k u_{ij}^{2}}
\label{eq8}
\end{equation}
\begin{equation}
gran\_error= \sum_{j=1}^n ||x_{j}-\hat{x_{j}}||^{2}
\label{eq9}
\end{equation}
\begin{equation}
PBM\_FVG=\left(\frac{1}{c} \cdot \frac{D_{c}}{\sqrt{gran\_error}}   \right)^{2}
\label{eq10}
\end{equation}
$\check{Z}$alik \cite{zalik2011221} proposed a new validity index based on the separation between clusters and the concept of clusters overlap instead of the compactness measure.
\begin{equation}
OS=\frac{\sum_{i=1}^n{\sum_{x_{j}\in C_{i}} O_{x_{j}}}}{\sum_{i=1}^k min_{j=1..k,i\neq j} ||v_{i}-{v_{j}}||}
\label{eq11}
\end{equation}
with:
\begin{equation}
O_{x_{j}} =\left\lbrace   
\begin{array}{ccc}
\frac{a}{b}&if&\frac{b-a}{b+a}<0.4 \\
0&otherwise
\end{array}
\right.
\end{equation}
\begin{equation}
a=\frac{1}{\left|c_{i}\right|} \sum_{x_{l} \in C_{i}} ||x_{j}-x_{l}||
\end{equation}
\begin{equation}
b=\frac{1}{\left|c_{i}\right|} \sum_{x_{l} \notin C_{i}} ||x_{j}-x_{l}||
\end{equation}
with $\left|c_{i}\right|$ is the cardinality of the $i^{th}$ cluster.
This CVI doesn't involve membership values. It is based on distance computation between data objects for the overlap measure and the distance between cluster centers as separation measure. The best partition is the one which has low intercluster overlap and high separation thus low value of $OS$. \vspace{-0.4cm} 
\section{New cluster validity index}
\subsection{Motivation}
The classical separation measure is based on computing distances between cluster centers. However, such measure has drawbacks. In Fig. 1, three clusters are generated from  Gaussian distributions. Cluster A and B are well separated. The distance between their centers is $d_{AB}=14.5$. Meanwhile, an overlap between cluster B and C is present but the distance between their centers is  $d_{BC}=13.2$. This case demonstrates the shortcoming of the separation measure based on distances between cluster centers.  
\subsection{Cluster validity index based on Jeffrey divergence}
The proposed validity index uses a separation measure based on the computation of Jeffrey divergence. 
The Jeffrey divergence between two distributions $P$ and $Q$ is given by:
\begin{equation}
JD(P,Q)=\sum_{x}\left(P(x)-Q(x)\right)ln\left(\frac{P(x)}{Q(x)}\right)
\end{equation}
JD is computed as the product of two terms. $(P(x)- Q(x))$ is proportional to the distance between two probability densities. $ln(\frac{P(x)}{Q(x)})$ which is also proportional to the level of separation between probability densities.
The closer $P(x)$ and $Q(x)$ to each other, the higher the overlap is and the lesser the value of Jeffrey divergence is.
Divergence allows to evaluate the extent to which two probability density functions  (PDF) differentiate.  Jeffrey divergence, unlike the Kullback-Leiber, is symmetric which reduces computational time in addition to being widely used in pattern recognition and computer vision applications \cite{4773151,6203369}. It is also numerically stable and robust with respect to noise and the size of the bins \cite{609331}. In our experiments, after applying a clustering algorithm, the density of each cluster is computed using PDF estimation techniques, then Jeffrey divergence is computed between pairs of clusters. Finally, the separation measure is determined.
\begin{figure}[h]
\centering
\includegraphics[scale=.37]{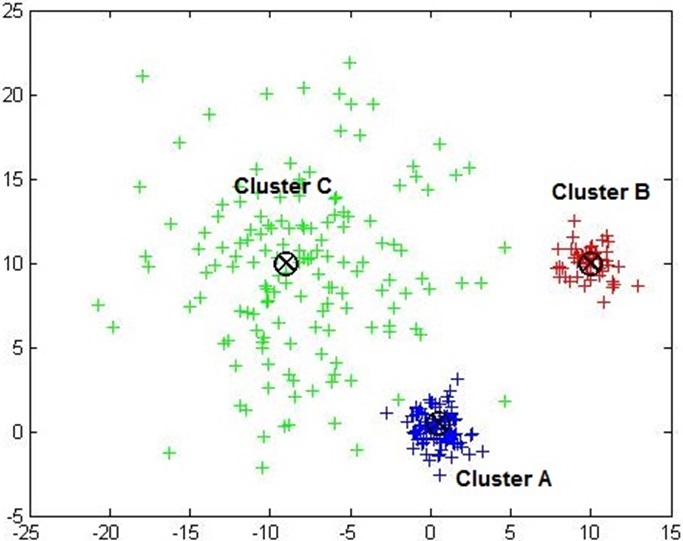}
\caption{Three clusters generated from Gaussian distributions. Cluster B and C are equally distant to cluster A but cluster C overlaps more with cluster A}
\label{fig:untitled2}
\end{figure}
\subsubsection{Cluster density estimation}
Let $x_{1},x_{2},\ldots, x_{N_{i}}$  be the $N_{i}$ features of the $i^{th}$ cluster $C_{i}$.\newline
We assume first that data are generated from multivariate Gaussian distribution given by:
\begin{equation}
p(x|\mu,\Sigma)=\frac{1}{(2\pi)^{d/2}|\Sigma|^{1/2}}exp\left(-\frac{1}{2}(x-\mu)^{T}\Sigma^{-1}(x-\mu)\right)
\end{equation}
Where $\mu$ and $\Sigma$ are the mean vector and covariance matrix of the data respectively. T is the transpose operator and $|$  $|$ is the determinant operator.
In order to determine the PDF, we need to estimate  parameters ($\mu$,$\Sigma$). We apply a maximum likelihood estimation \cite{citeulike:2180653} and we obtain the following estimations (see Annex A):
\begin{equation}
\hat{\mu}=\frac{1}{N_{i}}\sum_{i=1}^{N_{i}}x_{i}
\end{equation}
\begin{equation}
\hat{\Sigma}=\sum_{i=1}^{N_{i}}(x_{i}-\hat{\mu})^{T}(x_{i}-\hat{\mu})
\end{equation}
The Jeffrey divergence  between two clusters $C_{i}$ and $C_{j}$ generated from Gaussian distributions $N(\mu_{i},\Sigma_{i})$
and $N(\mu_{j},\Sigma_{j})$ is:
\begin{equation}
\begin{tabular}{  c l }
\(JD(C_{i},C_{j}=\frac{1}{2}\left( tr(\Sigma_{i}^{-1}\Sigma_{j})+tr(\Sigma_{j}^{-1}\Sigma_{i}) \right) + \)\\
\( \frac{1}{2} \left((\mu_{i}-\mu_{j})^{T}(\Sigma_{i}^{-1}+\Sigma_{j}^{-1})(\mu_{i}-\mu_{j})\right)-d \)
\end{tabular}
\label{jd}
\end{equation}
\newline
To generalize, let us assume that data are generated from any distribution. To estimate the PDF, we rely on the multivariate density distribution using the Gaussian kernel and define the density function as:
\begin{equation}
\hat{p}_{H}(x)=\frac{1}{N_{i}}\sum_{i=1}^ {N_{i}}K_{H}(x-x_{i})
\end{equation}
Where $H$ is the bandwidth and $K$ is a kernel function: 
\begin{equation}
K_{H}(x)=|H|^{-1/2}K(H^{-1/2}x)
\end{equation}
We use a Gaussian kernel for the density estimation:
\begin{equation}
K(x)=(2\pi)^{d/2}exp(-\frac{1}{2}x^{T}x)
\end{equation}
\subsubsection{Separation measure}
Based on the computation of Jeffrey divergence, we set up the separation measure as follows: after applying the clustering algorithm, we estimate the density of each cluster $C_{i}$. Divergence between $C_{i}$ and $C_{j,j=,1\ldots,k,j\neq i}$ is computed. After that, we take the minimum of calculated divergences. The separation measure is:
\begin{equation}
S=\sum_{i=1}^{k} Sep_{i}
\end{equation}
With:
\begin{equation}
Sep_{i}=\underset{j=1\ldots k,j\neq i}{min}(JD(C_{i},C_{j}))
\end{equation}
It is the sum of the minimum of  overlap, e.g Jeffrey divergence, between each cluster and other clusters. By choosing the minimum value, we are taking the least overlap degree for each cluster which is an indicator of its separability.
A high separation value is an indicator of a good partition.
\subsubsection{Compactness measure}
The compactness measure is computed as  the summation for all clusters  of the squared maximal distance of a feature $x_{j}$ belonging to the $i^{th}$ cluster $C_{i}$ to its center $v_{i}$. We denote the compactness measure $V$.
\begin{equation}
V=\sum_{i=1}^{k} \underset{x_{j}\in C_{i}}{max} \left\| x_{j}-v_{i} \right\|^{2}
\end{equation}
A low compactness measure indicates a good partition.
\subsubsection{Proposed cluster validity index: $I$}
The proposed cluster validity index $I$ is the ratio of the proposed separation measure to the compactness measure.
\begin{equation}
I=\frac{V}{S}
\end{equation}
A good partition is characterized by a high separation value and a low compactness. Thus, a low value of $I$ is an indicator of a good partition. \vspace{-0.2cm} 
\section{Experimental results}
In order to evaluate the performance of the proposed CVI, we conduct experiments on several synthetic and real data sets with known number of clusters. The classic evaluation methodology consists of applying a clustering algorithm such as FCM on each data set while varying the number of clusters and computing CVIs for each partition. The best CVI determines that the real number of clusters is the one that best fits the data. Figures 7-14 show the variation of each index according to the number of clusters.
To evaluate the performances of CVI, we are interested exclusively in checking if the maximum or minimum of each CVI coincides with the optimal number of clusters. The other values of each index are not relevant to this evaluation so curves are scaled for display purposes and the vertical axis has no relevant significance.
In addition, for the real data sets, we use a second evaluation method \cite{Gurrutxaga2011505}. In fact, the classic methodology is based on the assumption of the perfection of the clustering algorithm which is not always true. In the alternative methodology, the definition of the best partition is different. Indeed, the best partition is the one which is most similar to the perfect partition. Similarity is quantified using similarity measures such as Rand or adjusted Rand index. The best CVI is the index which determines that the most similar partition to the perfect partition is the best one. \newline We have computed the proposed measure for clusters of Fig.1; the separation measure between cluster A and B is $Sep_{AB}=69.7$ and between A and C is $Sep_{AC}=22.9$. We can see that the proposed measure reflects better the degree of separation between the three clusters. \vspace{-0.35cm}
\subsection{Synthetic data sets}  
We use four 2-dimensional data sets $S_{1},S_{2},S_{3}$ and $S_{4}$ \cite{Franti2006761} consisting of 5000 data vectors generated from   15 Gaussian clusters. These data sets are characterized by clusters of different shapes: circular and elliptical. In addition, clusters in data set $S_{1}$ are well separated compared to the other data sets where clusters overlap more and more until  becoming indistinguishable in data set $S_{4}$. Furthermore, we use a specially shaped set $R15$ data set \cite{1033218}. $R15$ consists of 2D 15 Gaussian clusters. Data sets are shown in Figures 2,3, 4, 5 and 6. \newline We apply FCM with a number of clusters varying from $k=10$ to $k=20$. The CVI which identifies $k=15$ as the best number of cluster is the one that outperforms the others.  Results for each set are represented in Fig. 7, 8, 9, 10 and 11. The best result for each CVI is marked in black. \newline
The findings demonstrate that the proposed validity index successfully determines the correct number of clusters for all data sets. On the other hand, indexes such as $PC$ and $PE$ are able to determine the correct number of clusters where there is a low overlapping between clusters such as in data set $S_{1}$. But we can notice that there is a slight variation in their values as overlapping degree gets higher unlike the proposed index $I$ and $XB$ index where discrimination between their values for each number of cluster is obvious. Note that the remaining indexes such PBM and PBMF\_FVG completely fail to determine the correct number of clusters in all cases and the best result is always found to be the lowest cluster number.
\begin{figure}[!h]
\centering
\includegraphics[scale=.35]{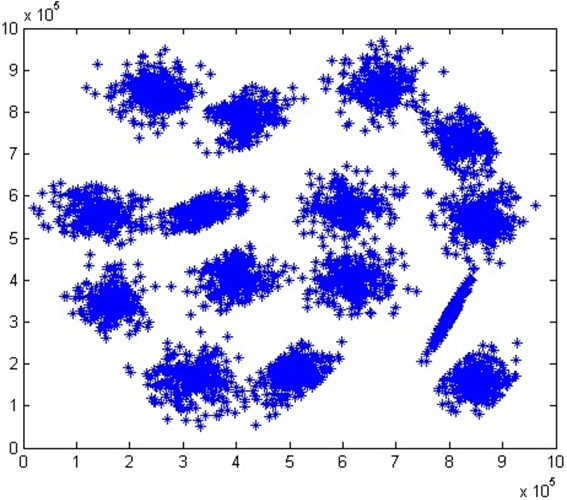}
\caption{Data set S1 with 15 Gaussian clusters}
\label{fig:untitled3}
\end{figure}
\begin{figure}[!h]
\centering
\includegraphics[scale=.37]{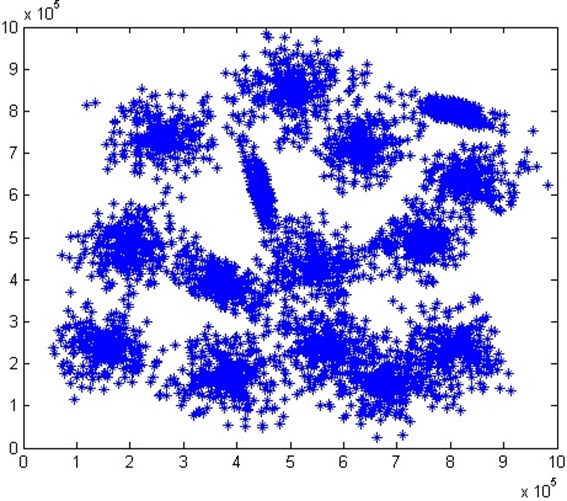}
\caption{Data set S2 with 15 Gaussian clusters}
\label{fig:untitled4}
\end{figure}
\vspace{-0.6cm}
\begin{figure}[!h]
\centering
\includegraphics[scale=.35]{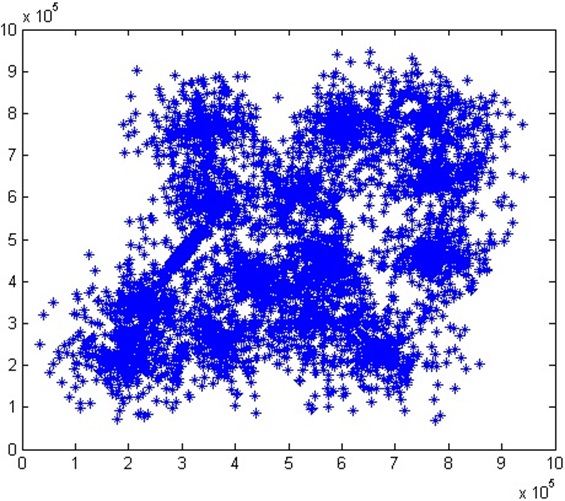}
\caption{Data set S3 with 15 Gaussian clusters}
\label{fig:untitled5}
\end{figure}
\begin{figure}[!h]
\centering
\includegraphics[scale=.35]{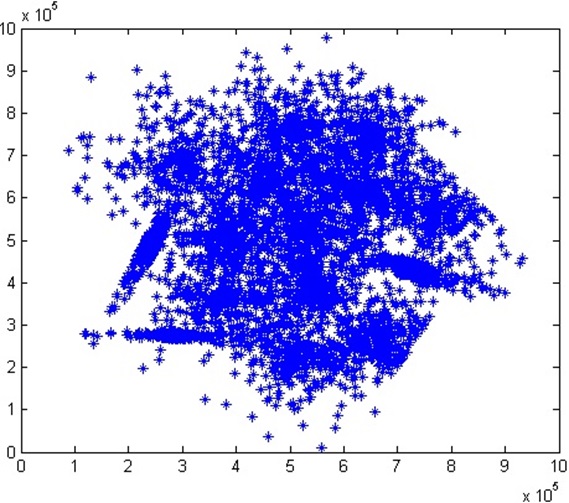}
\caption{Data set S4 with 15 Gaussian clusters}
\label{fig:untitled6}
\end{figure}
\vspace{-0.1cm}
\begin{figure}[!h]
\centering
\includegraphics[scale=.35]{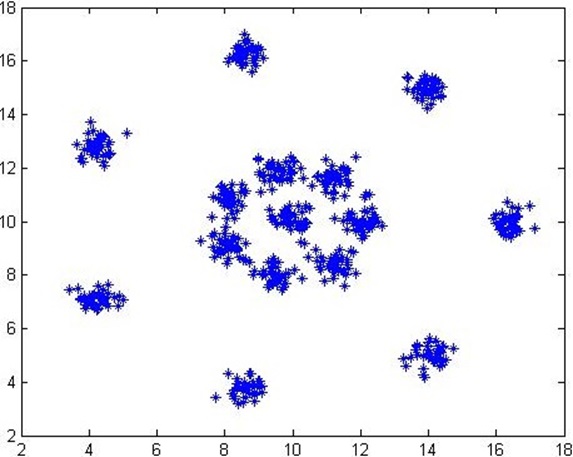}
\caption{Data set R15 with 15 Gaussian clusters}
\label{fig:untitled6}
\end{figure}

\vspace{-0.4cm}
\subsection{Real data set}
\subsubsection{Data}
\begin{itemize}
\renewcommand{\labelitemi}{$\bullet$}
\item \textbf{Balance data set} \cite{uci}: This data  set represents psychological experimental results. It consists of 625 4D features divided into 3 clusters. Each feature is classified as having balance to the right or to the left or being balanced. The attributes are the left weight, left distance, right weight and right distance. 
\item \textbf{Banana data set}\footnote{http://ida.first.fraunhofer.de/projects/bench/benchmarks.htm.}: This data set is banana shaped. It consists of 5300 features divided into two classes.
\item \textbf{Iris data set} \cite{uci}: Iris is a 4D data set  consisting of 150 features divided equally into 3 clusters: Iris setosa, Iris virginica and Iris versicolor. Two of these clusters heavily overlap. The 4D represents sepal length, sepal width, petal length and petal width. 
\item \textbf{ Extended Yale Database} \cite{GeBeKr01,KCLee05}: This database consists of 16128 images of 28 human subjects under 9 poses and 64 illumination conditions. In our experiments, we use the first 5 and 7 classes data with 64 images each denoted Yale\_5 and Yale\_7 respectively. We resize the cropped images into $24 \times 21$ then project the data onto a lower dimensional subspace using PCA.
\item \textbf{Hopkins 155}\footnote{http://www.vision.jhu.edu/data/hopkins155/}: This data set consists of 156 sequences of two and three motions which can be devided into three categories: checkerkboard, traffic and articulated sequences. Each sequence is a segmentation task. The checkerboard category consists of 104 sequences of indoor scenes taken with a handheld camera under controlled conditions. The checkerboard pattern on the objects is used to assure a large number of tracked points. Traffic category consists of 38 sequences of outdoor traffic scenes taken by a moving handheld camera. Articulated sequences display motions constrained by joints, head and face motions, people walking, etc.
\end{itemize}
\subsubsection{Classic methodology: results}
Results in Fig. 12 for Balance data set demonstrate that only the proposed index and the $P$ index are able to determine the optimal number of clusters. The behavior of indexes such as PC and PBM\_FVG seen with synthetic data sets is confirmed with the real data set, i.e the lowest number of clusters is chosen as the optimal number. These indexes represent a monotonic tendency with the number of clusters. \newline
In Fig. 13 for Banana data set, the proposed validity index estimated that the real number of cluster e.g $c^{*}=2$ is the optimal number of cluster. A large variation in index $I$ is noticed. This gives an information about the correct number of cluster. Almost all the validity indexes were able to estimate $c^{*}$. The monotonic tendency of some indexes such as $PE$ and $PC$ is also confirmed.\newline
The presence of overlap in the Iris data set makes it difficult to estimate the optimal number of cluster. Only index $I$ was able to determine $c^{*}=3$ as the optimal number of clusters as shown in Fig. 14. $PC$ and $PE$ always exhibit the monotonic tendency for the number of clusters. $PBM$ and $PBMF\_FVG$ showed the same behavior as for the previous data sets. \newline With eYale database, index I exhibits good performance. Indeed, for Yale\_5 data set as showeed in Fig. 15, only the proposed index was able to predict the real number of cluster as well as for Yale\_7 data set according to the results in Fig. 16.
\subsubsection{Alternative methodology: results}
We use four similarity measures namely the Rand index, the Fowlkes-Mallows (FM), the Jaccard (Jacc) and the adjsuted Rand (ARI) indexes \cite{jain} which are described
in Appendix C. We run FCM 100 times on the selected data and compute how many times each CVI determines the most similar partition as the best partition. Results for Balance, Banana and Iris are shown in tables 1, 2 and 3. \newline For the balance data set, index $I$ presents good performance. In fact, it is able to detect the best partition for all the similarity measures unlike other indexes which perform well with some similarity measures but fail with others such as PC and PE known for their monotonic tendency. \newline For the banana data set, the proposed index also presents good performances. It was able to determine the optimal partition with all similarity measures except Rand. \newline For Iris data set, only the proposed index $I$ was able to determine the best partition in case of Rand, Jaccard and adjusted Rand indexes. Such result goes with the result obtained for Iris data in the classic methodology. \newline For the Hopkins 155 data, we use the 1R2RCT\_A, 1R2RCT\_B and car5 data sets which have three motions each. A classic preprocessing step before using these data is to project it onto a subspace using PCA. In fact, it is known for these data that the feature trajectories of $n$ motions in a video almost perfectly lie in a $4n$ dimensional subspace. Thus, PCA projection will reduce the dimensionality with structure preservation \cite{6482137}. Results are shown in tables 4, 5 and 6. The overall performance of the proposed index is better than the other indexes for 1R2RCT\_A for all the similarity measures except ARI. Other indexes completely fail with Rand, FM and Jacc measures. With 1R2RT\_B, index I presents also good performance with a success of 100\% for Rand, FM and Jacc. PC and XB in this case present also the same performance. OSV and P perform well only with ARI. Results for car5 demonstrate also the outperformance of index I with Rand FM and ARI. In general, results on Hopkins 155 demonstrate that the proposed index I presents the best success rate compared to other indexes. 
\vspace{-0.3cm}
\subsection{Discussion}
The proposed validity index is highly useful especially for data sets where overlap between clusters is present. While some indexes are robust or completely fail in the presence of a low degree of overlapping, proposed index $I$ is robust and able to estimate the optimal number of cluster even when clusters highly overlap such as in data sets  $Iris$ and $S_{4}$ data sets where all or some clusters are hardly distinguishable. The design of the separation measure based on the density of clusters is an indicator of the degree of overlap between clusters as shown in section 4 which couldn't be determined with the classical separation measure based on the distance between cluster centers. However, in case of high dimensional data, the computation of the proposed cluster validity index may be computationally expensive. For example, in case of Gaussian clusters, the computation of Jeffrey divergence \eqref{jd} requires the inversion of the covariance matrix of size ($d\times d$). Using Cholesky decomposition, this operation has a complexity of $O(d^{3})$.
\vspace{-0.3cm}
\section{Conclusion}
Cluster validity indexes are used to compare the performances of clustering algorithms and determine the optimal number of clusters that best fits the data. Most of  CVIs available in the literature use a separation measure based on distance computation between cluster centers. It has been proven that such measure is not efficient especially in the case of overlapping clusters where it gives misleading results.\newline
In this work, we  proposed a new cluster validity index based on a new separation measure. The optimal partition is obtained with the lowest value of the index which means a high separation and low compactness. The proposed separation measure, unlike the classic ones, is based on density estimation of the obtained clusters. Jeffrey divergence is computed afterward informing us about the degree of overlap between pairs of clusters. The experimentation involved synthetic and real data sets. Comparisons between the proposed CVI and other CVIs have been conducted. Results demonstrated that the proposed CVI outperforms other indexes especially in the case of a data set containing overlapping clusters. In a future work, we will attempt to incorporate a local adaptive distance in the design of the CVI. This adaptive approach is expected to give significant separation and compactness measures.
\vspace{-0.3cm}
\section*{Aknowledgment}
This publication was made possible by NPRP grant \# 4-1165-
2-453 from the Qatar National Research Fund (a member of
Qatar Foundation). The statements made herein are solely the
responsibility of the authors.
\begin{figure}[!ht]
\centering
\includegraphics[scale=.6]{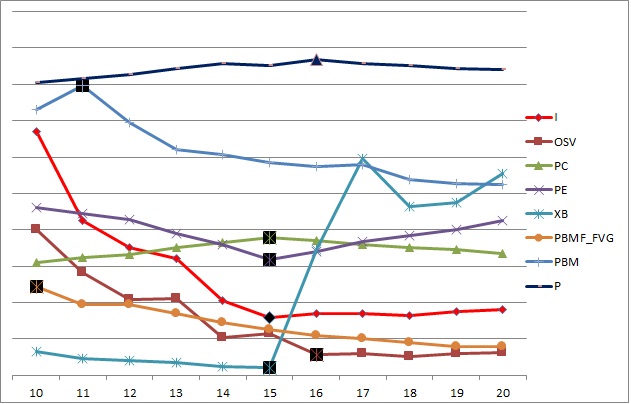}
\caption{CVIs for S1 with different number of clusters (lines are scaled for display purpose)}
\label{fig:untitled7}
\end{figure}
\begin{figure}[!ht]
\centering
\includegraphics[scale=.6]{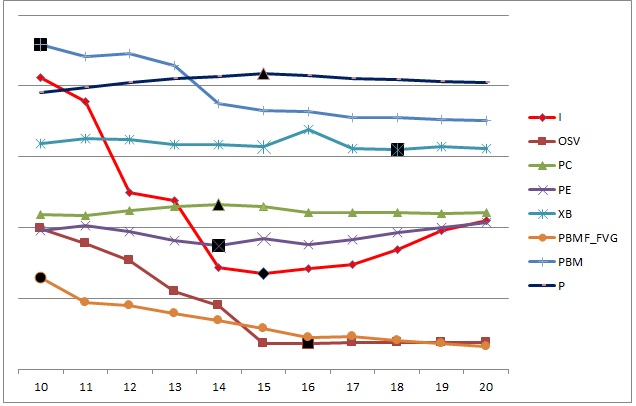}
\caption{CVIs for S2 with different number of clusters (lines are scaled for display purpose)}
\label{fig:untitled7}
\end{figure}
\begin{figure}[!ht]
\centering
\includegraphics[scale=.6]{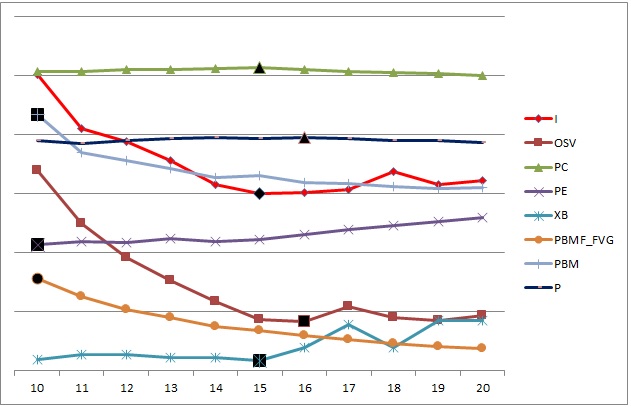}
\caption{CVIs for S3 with different number of clusters (lines are scaled for display purpose)}
\label{fig:untitled7}
\end{figure}
\begin{figure}[!ht]
\centering
\includegraphics[scale=.6]{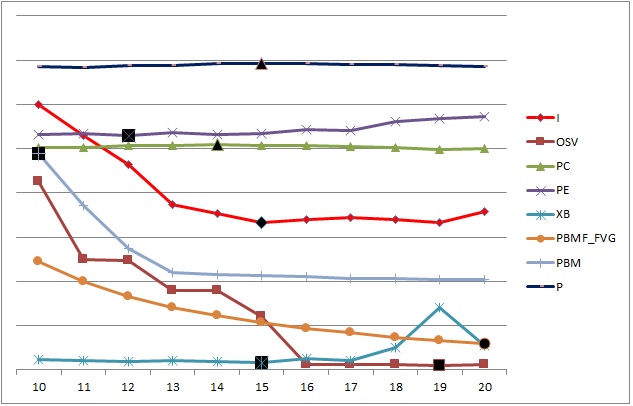}
\caption{CVIs for S4 with different number of clusters (lines are scaled for display purpose)}
\label{fig:untitled7}
\end{figure}
\begin{figure}[!ht]
\centering
\includegraphics[scale=.6]{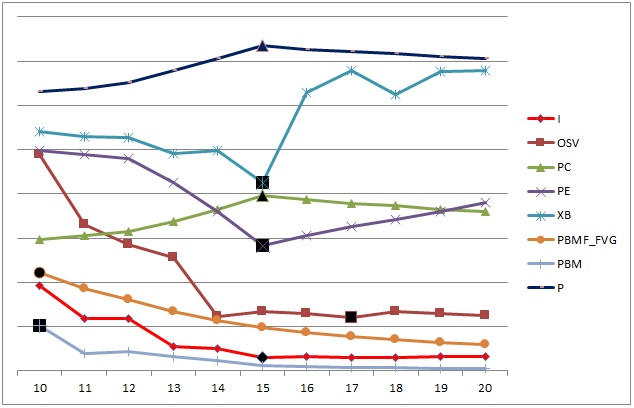}
\caption{CVIs for R15 with different number of clusters (lines are scaled for display purpose)}
\label{fig:untitled7}
\end{figure}
\begin{figure}[!ht]
\centering
\includegraphics[scale=.6]{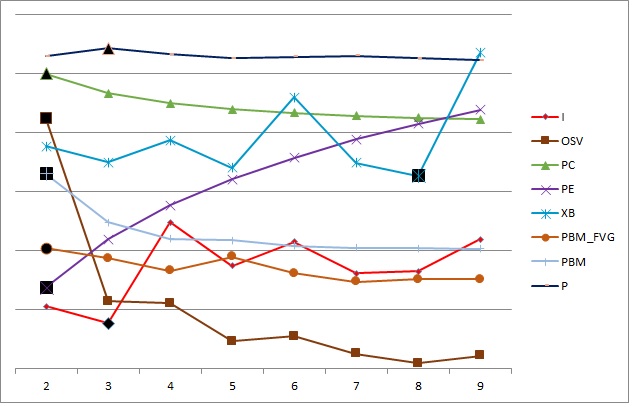}
\caption{CVIs for Balance data set with different number of clusters (lines are scaled for display purpose)}
\label{fig:untitled7}
\end{figure}
\newline
\begin{figure}[!ht]
\centering
\includegraphics[scale=.6]{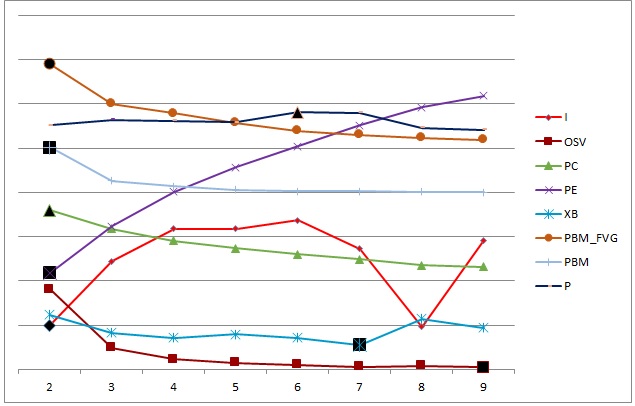}
\caption{CVIs for Banana data set with different number of clusters (lines are scaled for display purpose)}
\label{fig:untitled7}
\end{figure}
\begin{figure}[!ht]
\centering
\includegraphics[scale=.6]{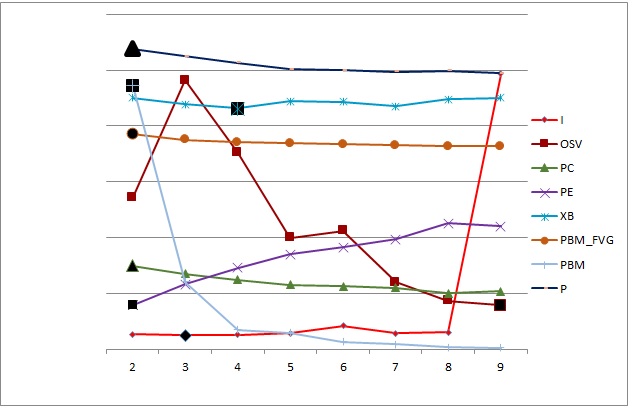}
\caption{CVIs for Iris data set with different number of clusters (lines are scaled for display purpose)}
\label{fig:untitled7}
\end{figure}
\begin{figure}[!ht]
\centering
\includegraphics[scale=.6]{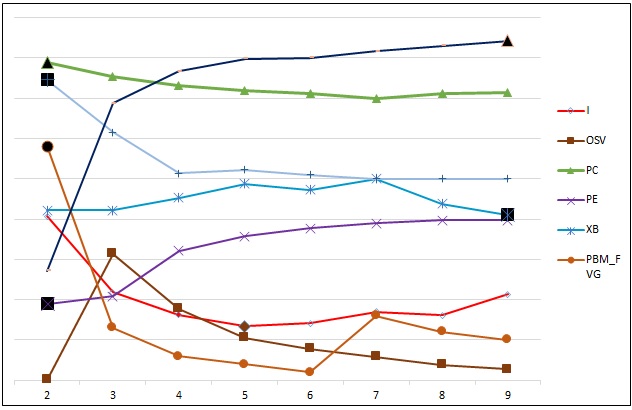}
\caption{CVIs for Yale\_5 data set with different number of clusters (lines are scaled for display purpose)}
\label{fig:untitled7}
\end{figure}
\begin{figure}[!ht]
\centering
\includegraphics[scale=.6]{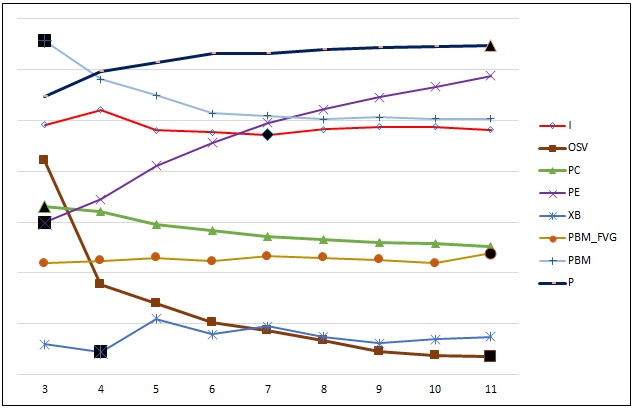}
\caption{CVIs for Yale\_7 with different number of clusters (lines are scaled for display purpose)}
\label{fig:untitled7}
\end{figure}
\newpage
\begin{table}[!ht]
\caption{Similarity measures for balance data set}
\begin{center}
\begin{tabular}{c c c c c c c c c}
		\hline
    Similarity  & PC & PE & P & XB & PBM & PBM& OSV &I  \\
			 measure &    &    &   &    &     & \_FVG& & \\
		\thickhline
   \centering Rand								& 0  & 0  & 0 & 0  &  0  &  0      & 75  &9 \\

		FM  								& 100& 100& 0 & 0  &100  &  100    & 0   &69 \\

		Jacc 								& 100& 100& 0 & 0  &100  &  100    & 0   &67 \\

		ARI  								& 0  & 0  & 0 & 0  &  0  &  0      & 79  &11 \\
		\hline	
\end{tabular}
\end{center}
\end{table}
\vspace{-1cm}
\begin{table}[!ht]
\caption{Similarity measures for banana data set}
\begin{center}
\begin{tabular}{ p{1cm} c c c c c c c c }
		\hline
    Similarity   & PC & PE & P & XB & PBM & PBM & OSV &I  \\
		 measure &    &    &   &    &     & \_FVG& & \\
		\thickhline
    Rand								& 0  & 81 & 8 & 58 &  0  &  0       & 29  &0 \\

		FM  								& 100& 100& 0 & 0  &100  &  100    & 0   &100 \\

		Jacc 								& 100& 100& 0 & 0  &100  &  100    & 0   &100 \\

		ARI  								& 100& 0  & 0 & 0  &  0  &  0      & 0   &100 \\
		\hline	
\end{tabular}
\end{center}
\end{table}
\vspace{-1cm}
\begin{table}[!ht]
\caption{Similarity measures for Iris data set}
\begin{center}
\begin{tabular}{c c c c c c c c c }
		\hline
    Similarity & PC & PE & P & XB & PBM & PBM& OSV &I  \\
		 measure &    &    &   &    &     & \_FVG& & \\
		\thickhline
    Rand								& 0  & 0  & 0 & 0  &  0  &  0      & 0    &17 \\
		
		FM  								& 100& 100& 0 & 100&100  &  100    & 0    &0 \\
		
		Jacc 								& 0  & 0  & 0 & 0  &0    &  0      & 0    &15 \\
		
		ARI  								& 0  & 0  & 0 & 0  &  0  &  0      & 0    &84 \\
		\hline	
\end{tabular}
\end{center}
\end{table}
\begin{table}[!ht]
\caption{Similarity measures for 1R2RCT\_A data set}
\begin{center}
\begin{tabular}{c c c c c c c c c }
		\hline
    Similarity  & PC & PE & P & XB & PBM & PBM & OSV &I  \\
		 measure &    &    &   &    &     & \_FVG& & \\
		\thickhline
    Rand								& 0  & 0  & 0 & 0  &  0  &  0       & 0    &80 \\
		
		FM  								& 0  & 0  & 0 & 0  &  0  &  0       & 0    &80 \\
		
		Jacc 								& 0  & 0  & 0 & 0  &  0  &  0       & 0    &80 \\
		
		ARI  								& 100& 100& 0 & 0  &100  &  100     & 0    &0 \\
		\hline	
\end{tabular}
\end{center}
\end{table}
\begin{table}
\caption{Similarity measures for 1R2RT\_B data set}
\begin{center}
\begin{tabular}{c c c c c c c c c }
		\hline
    Similarity  & PC & PE & P & XB & PBM & PBM & OSV &I  \\
		 measure &    &    &   &    &     & \_FVG& & \\
		\thickhline
    Rand								& 100& 0  & 0 & 100&  0  &  0       & 0    &100 \\
		
		FM  								& 100& 0  & 0 & 100&  0  &  0       & 0    &100 \\
		
		Jacc 								& 100& 0  & 0 & 100&  0  &  0       & 0    &100 \\
		
		ARI  								& 0  &100 & 90& 0  &  0  &  0       & 66    &0 \\
		\hline	
\end{tabular}
\end{center}
\end{table}
\begin{table}[!ht]
\caption{Similarity measures for car5 data set}
\begin{center}
\begin{tabular}{c c c c c c c c c }
		\hline
    Similarity & PC & PE & P & XB & PBM & PBM & OSV &I  \\
		 measure &    &    &   &    &     & \_FVG& & \\
		 \thickhline
    Rand								& 0  & 0  & 0 & 76 &  0  &  0       & 0    &96 \\
		
		FM  								& 41 & 41 & 0 & 61 &  0  &  41      & 41    &82 \\
		
		Jacc 								& 99 & 99 & 0 & 0  &  0  &  99      & 99    &0 \\
		
		ARI  								& 0  & 0  & 0 & 0  &76   &  0       & 0    &96 \\
		\hline	
\end{tabular}
\newline
\end{center}
\end{table}
\section*{Annex A. Parameters estimation}
In all the expressions below, x is a vector of random variables whose mean vector and covariance matrix are given by: 
$E(x) = \mu $ and $ E((x-\mu)(x-\mu)^{T}) = \Sigma$ where $E$ means the expectation.
Using matrix properties:
\begin{itemize}
\item $\frac{\partial A^{T}\cdot x}{\partial x}=\frac{\partial x^{T}\cdot A}{\partial x}= A$
\item $\frac{\partial x^{T}\cdot A\cdot x}{\partial x}=A^{T}+A$
\item $\frac{\partial }{\partial A} log |A|= \left(A^{-1}\right)^{T}$
\item $\frac{\partial }{\partial A} tr \left[AB\right]=tr \left[BA\right]=B^{T}$
\end{itemize}
The log likelihood of the multivariate Gaussian distribution is given by:
\begin{equation}
\begin{tabular}{  c l }
\(L(x|\mu,\Sigma )=\frac{-nD}{2}log(2\pi) - \frac{n}{2}log(|\Sigma|)  - \)\\
\( \frac{1}{2} \sum_{i=1}^{n}(x_{i}-\mu)^{T} \Sigma^{-1}(x_{i}-\mu) \)
\end{tabular}
\label{jd}
\end{equation}
The estimates of the mean and covariance matrix are determined by computing the derivatives of $L(x|\mu,\Sigma )$ with relative to $\mu$ and $\Sigma$ and set it equal to zero.
\newline
\begin{equation*}
\begin{tabular}{ r c l }
  \(\frac{\partial L(x|\mu,\Sigma )}{\partial \mu}\) & \(=\) & \(\frac{\partial}{\partial \mu} \left(\sum_{n}^{i=1}(x_{i}-\mu)^{T}\Sigma^{-1}(x_{i}-\mu)\right)\) \\
		& \(\) &\\
   & \(=\) & \( \frac{\partial}{\partial \mu} ( \sum_{i=1}^{n} ( x_{i}^{T}\Sigma^{-1}x_{i} - \mu^{T}\Sigma^{-1}x_{i} - x_{i}^{T}\Sigma^{-1}\mu \) \\
	& \(\) &\\
	& \(\) & \(+ \mu^{T} \Sigma^{-1}\mu  ) )     \) \\
	& \(\) &\\
   & \(=\) & \( \sum_{i=1}^{n}\left( \Sigma^{-1}x_{i}+ \left(\Sigma^{-1} \right)^{T}x_{i}\right) - \) \\
	& \(\) &\\
	& \(\) & \(N\left(\Sigma^{-1}+\left(\Sigma^{-1} \right)^{T}\right)\mu\) \\
	& \(\) &\\
	& \(=\) &\(0\)\\
		& \(\) &\\
	 & \(\Rightarrow\) & \(\hat{\mu}=\frac{1}{n}\sum_{i=1}^{n}x_{i}\)\\
\end{tabular}
\end{equation*}
\newline
\begin{equation*}
\begin{tabular}{ r c l }
\( \frac{\partial L(x|\mu,\Sigma)}{\partial \Sigma^{-1}}\) & \(=\) &  \( \frac{\partial}{\partial \Sigma^{-1} } ( -\frac{N}{2}log(|\Sigma|) -\)\\
	& \(\) &\\
&\(\) &\(\frac{1}{2} \sum^{n}_{i=1} (x_{i}-\mu)^{T}\Sigma^{-1}(x_{i}-\mu) )\) \\
	& \(\) &\\
& \(\propto \) & \( \frac{\partial}{\partial \Sigma^{-1} } ( -\frac{N}{2}log(|\Sigma|) -\frac{1}{2} \sum^{n}_{i=1} tr[  \Sigma^{-1}  (x_{i}-\mu)\cdot \) \\
& \(\) &\\
& \(\) & \((x_{i}-\mu)^{T} ]    ) \) \\
	& \(\) &\\
& \(= \) & \( \frac{\partial}{\partial \Sigma^{-1} } ( \frac{N}{2}log(|\Sigma^{-1}|) -\frac{1}{2}  tr[ \sum^{n}_{i=1} \Sigma^{-1}  (x_{i}-\mu)\cdot \)\\
& \(\) &\\
 & \(\) &\( (x_{i}-\mu)^{T} ]    ) \)\\
	& \(\) &\\
& \(= \)&\(0 \) \\
	& \(\) &\\
& \(\Rightarrow\) & \( \hat{\Sigma}=\sum_{i=1}^{n}(x_{i}- \hat{\mu})(x_{i}- \hat{\mu})^{T} \)
\end{tabular}
\end{equation*}
\section*{Annex B. Jeffrey divergence for multivariate Gaussian distribution}
We use the following formula:
\begin{itemize}
\item $ E(x^{T}\cdot A \cdot x)= tr(A\cdot \Sigma)+ \mu^{T}\cdot A\cdot \mu$
\item $\left\langle \cdot \right\rangle$ is the expectation symbol.
\end{itemize}
\begin{equation*}
p(x|\mu_{1},\Sigma_{1})=\frac{1}{(2\pi)^{d/2}|\Sigma_{1}|^{1/2}}exp\left(-\frac{1}{2}(x-\mu_{1})^{T}\Sigma_{1}^{-1}(x-\mu_{1})\right)
\end{equation*}
\begin{equation*}
q(x|\mu_{2},\Sigma_{2})=\frac{1}{(2\pi)^{d/2}|\Sigma_{2}|^{1/2}}exp\left(-\frac{1}{2}(x-\mu_{2})^{T}\Sigma_{2}^{-1}(x-\mu_{2})\right)
\end{equation*}
\begin{equation*}
JD(p,q)= KL(p/q)+KL(q/p)
\end{equation*}
\newline
Where $KL$ is the Kullback-Leiber divergence.
\newline
\begin{equation*}
\begin{tabular}{ r c l }
\(KL(p/q)\)& \(=\) & \(\int log\left( p(x)-q(x)\right)p(x)\)\\
& \(\) &\\
& \(=\) & \(\int ( \frac{1}{2}log\left(\frac{|\Sigma_{2}|}{|\Sigma_{1}|}\right) - \frac{1}{2}(x-\mu_{1})^{T}\Sigma_{1}^{-1}(x-\mu_{1}) \)\\
& \(\) &\\
& \(\) & \(+\frac{1}{2}(x-\mu_{2})^{T}\Sigma_{2}^{-1}(x-\mu_{2})) p(x) \)\\
& \(\) &\\
& \(=\) & \( \frac{1}{2} log\left(\frac{|\Sigma_{2}|}{|\Sigma_{1}|}\right)- \frac{1}{2} \left\langle (x-\mu_{1})^{T}\Sigma_{1}^{-1}(x-\mu_{1})  \right\rangle \)\\
& \(\) &\\
& \(\) & \(+ \frac{1}{2}\left\langle (x-\mu_{2})^{T}\Sigma_{2}^{-1}(x-\mu_{2})  \right\rangle  \)\\
& \(\) &\\
& \(=\) &\( \frac{1}{2} \left( log\left(\frac{|\Sigma_{2}|}{|\Sigma_{1}|}\right) - tr \left(\Sigma_{1}^{-1}\Sigma_{1}\right) + tr \left(\Sigma_{2}^{-1}\Sigma_{1}\right)\right) \)\\
& \(\) &\\
& \(\) & \(+ \frac{1}{2} \left( (\mu_{1}-\mu_{2})^{T}\Sigma_{2}^{-1}(\mu_{1}-\mu_{2})\right) \)\\
& \(\) &\\
& \(=\) & \( \frac{1}{2} \left( log\left(\frac{|\Sigma_{2}|}{|\Sigma_{1}|}\right) -d + + tr \left(\Sigma_{2}^{-1}\Sigma_{1}\right)\right)\)\\
& \(\) &\\
& \(\) &\( + \frac{1}{2} \left( (\mu_{1}-\mu_{2})^{T}\Sigma_{2}^{-1}(\mu_{1}-\mu_{2})  \right) \)
\end{tabular}
\end{equation*}
Thus:
\begin{equation*}
\begin{tabular}{ r c l }
\(JD(p,q)\)& \(=\) & \( \frac{1}{2}\left( tr(\Sigma_{1}^{-1}\Sigma_{2})+tr(\Sigma_{2}^{-1}\Sigma_{1}) \right) + \)\\
& \(\) & \( \frac{1}{2} \left((\mu_{1}-\mu_{2})^{T}(\Sigma_{1}^{-1}+\Sigma_{2}^{-1})(\mu_{1}-\mu_{2})\right)-d \)
\end{tabular}
\end{equation*}
\section*{Annex C. Similarity measures}
The four similarity measures we have used are described in this section. Let $P1$ and $P2$ two partitions. We define $a$ as the number of object pairs that belong to the same clusters in both $P1$ and $P2$. Let $b$ be the number of object pairs that belongs to different clusters in both pairs. Let $c$ be the number of object pairs that belong to the same clusters in $P1$ but in different clusters in $P2$. Finally, let $d$ the number of pairs that belong to different clusters in $P1$ but belong to the same cluster in $P2$. The four similarity measures are defined as:
\begin{equation*}
\begin{tabular}{ r c l }
\(Rand\)&\(=\)&\(\frac{a+b}{a+b+c+d}\) \\
\end{tabular}
\end{equation*}
\begin{equation*}
\begin{tabular}{ r c l }
\(Fowlkes-Mallows\) & \(=\) & \( \frac{a}{\sqrt{(a+d)(a+c)}} \) \\
\end{tabular}
\end{equation*}
\begin{equation*}
\begin{tabular}{ r c l }
\(Jaccard\)& \(=\)&\(\frac{a}{a+c+d}\)\\\
\end{tabular}
\end{equation*}
\begin{equation*}
\begin{tabular}{ r c l }
\(Adjsuted-Rand\)&\(=\)&\(\frac{a-\frac{(a+d)(a+c)}{a+b+c+d}}{\frac{a+b+c+d}{2}-\frac{(a+d)(a+c)}{a+b+c+d}}\)\\
\end{tabular}
\end{equation*}


\end{document}